\documentclass{article}

     \PassOptionsToPackage{numbers, compress}{natbib}

     \usepackage[preprint]{neurips_2019}

\usepackage[utf8]{inputenc} 
\usepackage[T1]{fontenc}    
\usepackage{hyperref}       
\usepackage{url}            
\usepackage{booktabs}       
\usepackage{amsfonts}       
\usepackage{nicefrac}       
\usepackage{microtype}      
\usepackage{enumitem}
\usepackage{graphicx}
\usepackage{tabularx}
\usepackage{adjustbox}
\usepackage{threeparttable}
\usepackage{subfig}
\usepackage{amsmath}
\usepackage{algorithm}
\usepackage{algorithmic}
\usepackage{multirow}
\usepackage{tabu}
\usepackage{amssymb}
\DeclareCaptionLabelFormat{singleparen}{#2}
\newcommand{\figref}[1]{Fig.~\ref{#1}}
\newcommand{\tabref}[1]{Table~\ref{#1}}

\newcommand{\secref}[1]{Sec.~\ref{#1}}
\newcommand{\ie}{\textit{i.e.}}
\newcommand{\eg}{\textit{e.g.}}
\newcommand{\etal}{\textit{et.al.}}

\hypersetup{
     colorlinks   = true,
     citecolor    = green
}
\hypersetup{linkcolor=red}

\title{Align-and-Attend Network for Globally and Locally Coherent Video Inpainting}
\author{
  Sanghyun Woo \\
  EE, KAIST \\
  Daejeon, Korea \\
  \texttt{shwoo93@kaist.ac.kr} \\
  \And
  Dahun Kim \\
  EE, KAIST \\
  Daejeon, Korea \\
  \texttt{mcahny@kaist.ac.kr} \\
  \And
  KwanYong Park \\
  EE, KAIST \\
  Daejeon, Korea \\
  \texttt{pkyong7@kaist.ac.kr} \\
  \And
  Joon-Young Lee \\
  Adobe Research \\
  San Jose, CA, USA\\
  \texttt{jolee@adobe.com} \\
  \And
  In So Kweon \\
  EE, KAIST \\
  Daejeon, Korea \\
  \texttt{iskweon@kaist.ac.kr} \\
}
\begin{document}
\maketitle


\begin{abstract}
We propose a novel feed-forward network for video inpainting. We use a set of sampled video frames as the reference to take visible contents to fill the hole of a target frame. Our video inpainting network consists of two stages. The first stage is an alignment module that uses computed homographies between the reference frames and the target frame. The visible patches are then aggregated based on the frame similarity to fill in the target holes roughly. The second stage is a non-local attention module that matches the generated patches with known reference patches (in space and time) to refine the previous global alignment stage. Both stages consist of large spatial-temporal window size for the reference and thus enable modeling long-range correlations between distant information and the hole regions. Therefore, even challenging scenes with large or slowly moving holes can be handled, which have been hardly modeled by existing flow-based approach. Our network is also designed with a recurrent propagation stream to encourage temporal consistency in video results. Experiments on video object removal demonstrate that our method inpaints the holes with globally and locally coherent contents.
\end{abstract}

\section{Introduction}

Video inpainting aims to fill spatial-temporal holes with plausible content in a video. It is a practical and crucial problem as it could be beneficial for various video editing and restoration tasks. However, it is very challenging to maintain both spatial and temporal consistency; the inpainted contents must be spatially plausible, and temporally coherent at the same time.

Early works for video inpainting use a patch-based optimization technique~\cite{wexler2004space,granados2012not,newson2014video,huang2016temporally}. Among them, Huang \etal~\cite{huang2016temporally} proposed a global flow field based optimization to preserve the temporal consistency, and show the state-of-the-art quality video results. However, the trade-off against the effectiveness is its limited practicality due to its intensive computational cost and vulnerability to noisy optical flows.  Recently, two seminar works have proposed deep feed-forward methods for video inpainting~\cite{wang2018video2,kim2019video}. Wang \etal~\cite{wang2018video2} proposed CombCN by combining 3D and 2D CNNs, but their setting works on low-resolution videos with fixed square holes, limiting its application to general video object removal. Kim \etal~\cite{kim2019video} proposed VINet which aggregates information by flow warping from nearby frames to the target frame. However, its internal dependency on the optical flow restricts the size of temporal search window, which sometimes leads to boundary artifacts and blurry textures inconsistent with global video contents.

\begin{figure*}[t]
\centering
\includegraphics[width=0.85\textwidth]{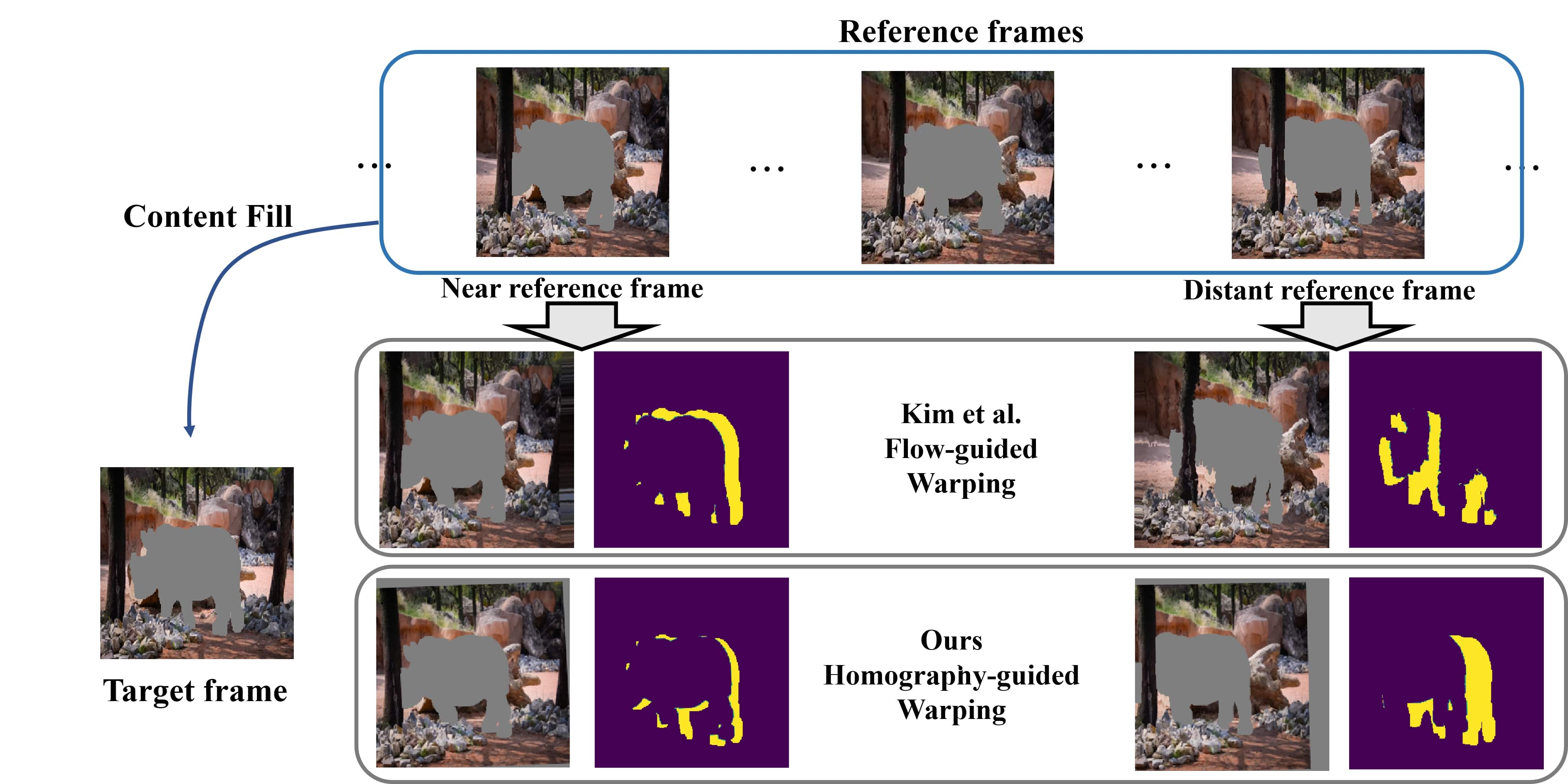}
\caption{\textbf{Comparison of two different warping methods.} The purple-yellow mask shows the region that can be filled up in the target frame.}
\label{fig:concept_overview}
\end{figure*}

To overcome the aforementioned issues, we propose a novel coarse-to-fine network for video inpainting. We use a set of sampled video frames as the reference to take visible contents to fill the hole of a target frame. Our proposed network consists of two stages. The first stage is an alignment module that uses computed homographies between the reference frames and the target frame. Despite being able to model only global transformations (\eg, affine, perspective), homography based alignment provides much larger temporal search window compared to the optical flow based counterpart,\eg,~\cite{kim2019video}, as illustrated in~\figref{fig:concept_overview}. The visible patches are then aggregated to roughly fill in the target holes. The second stage consists of a non-local attention module that matches the generated patches with known reference patches in space and time, and a softmax that temporally pools the most relevant patches. This refinement stage compensates real motions that cannot be modeled by previous global transformations. Both stages consist in large spatial-temporal window size for the reference, and thus enable modeling long-range correlations between distant information and the hole regions. Therefore, even challenging scenes with large or slowly moving holes can be handled. Our network is also designed with a decoder to synthesize the contents that are never visible throughout the video, and a recurrent propagation stream to encourage temporal consistency in video results.

We show that our video results are more semantically plausible, and temporally smooth compared to the previous methods. Our model sequentially processes video frames of arbitrary length and runs at a near real-time rate.

\section{Related Work}

\textbf{Traditional inpainting methods.}
Early works for image inpainting can broadly fall into either diffusion-based~\cite{ballester2001filling,bertalmio2000image,levin2003learning} or patch-based methods~\cite{bertalmio2003simultaneous,darabi2012image,efros2001image,simakov2008summarizing}. The former propagates texture from the hole boundaries towards the hole center, and works well with small holes, but suffers artifacts and noisy results with large holes. The latter tries to match and copy the nearest neighbor background patches, and is widely deployed in practical applications.

For videos, Granados \etal~\cite{granados2012background} and Newson \etal~\cite{newson2014video} proposed to align the frames in addition to using the optical flow or 3D PatchMatch search. Huang \etal~\cite{huang2016temporally} jointly optimize global flow and colors throughout a video for long-term temporal consistency. As mentioned earlier, these methods are heavy in computation time, prone to flow errors, and not able to capture high-level semantics.



\textbf{Learning-based inpainting methods.}
Deep learning based methods have achieved great success on the image inpainting task~\cite{pathak2016context,iizuka2017globally,yu2018generative,liu2018image,yu2018free}. They proposed to use Convolutional Neural Network together with Generative Adversarial Networks~\cite{pathak2016context}, global and local discriminators to improve spatial coherency~\cite{iizuka2017globally}, a coarse-to-fine model with contextual attention~\cite{yu2018generative}, partial convolution~\cite{liu2018image} and gated convolution~\cite{yu2018free} to handle free-form masks. However, they do not consider any consistencies between frames when applied to videos. 


Recently, two deep learning based Methods have been proposed for video inpainting task~\cite{wang2018video2,kim2019video}. CombCN~\cite{wang2018video2} has a 3D CNN with following 2D CNNs, where the 3D CNN part captures the temporal structure from low-resolution video. VINet~\cite{kim2019video} deals with real video object removal task by collecting visible information from nearby frames via flow warping. Nevertheless, their fundamental limitation is on their small spatial-temporal window size, which limits their performances for scenes with large and slowly moving holes.


\section{Proposed Algorithm}

Let $X_1^T := \{X_1, X_2, ..., X_T\}$ be a set of video frames with spatial-temporal holes, and   $Y_1^T := \{Y_1, Y_2, ..., Y_T\}$ be the reconstructed ground truth frames. We aim to learn mappings $G : X_1^T \rightarrow Y_1^T$, such that the prediction $\hat{Y}_1^T$ be as close as possible to the ground truth video $Y_1^T$,  while being plausible and consistent in space and time. This can also be formulated as a conditional video generation task~\cite{wang2018video,lee2019instert} where we estimate the conditional $p({Y}_1^T|X_1^T)$. To simplify the problem, we base on a Markov assumption~\cite{wang2018video,kim2019video} to factorize the conditional into a product form, such that the generation of the $t$-th target frame $\hat{Y}_t$ is dependent on 1) current input frame $X_t$, 2) two previous output frames $\hat{Y}_{t-2}^{t-1}$, and 3) a set of sampled reference frames $R$ as:

\begin{equation}
    p(\hat{Y_1^T}|X_1^T) = \prod_{t=1}^{T} p(\hat{Y}_t|X_{t}, \hat{Y}_{t-2}^{t-1}, R).
\label{eqn:our_formulation7}
\end{equation}

The main idea of our approach is to use a set of sampled reference frames, $R$, that contains sufficiently large temporal search window, so that the visible information in the window can be fetched to inpaint the target frame with the globally coherent contents. According to our preliminary experiments, we sample every 10-th frames in a video to construct $R$, and this provides much larger temporal window size than previous approaches~\cite{wang2018video2,kim2019video}. Another important design is to enforce each prediction to be temporally coherent with the past predictions,$\hat{Y}_{t-2}^{t-1}$ . With this recurrent pathway, our model runs in an auto-regressive manner. Our proposed method outperforms existing learning-based methods~\cite{wang2018video2,kim2019video}, and performs on par with the optimization-based method~\cite{huang2016temporally} while running at much faster speed.

\begin{figure*}[t]
\centering
\includegraphics[width=1\textwidth]{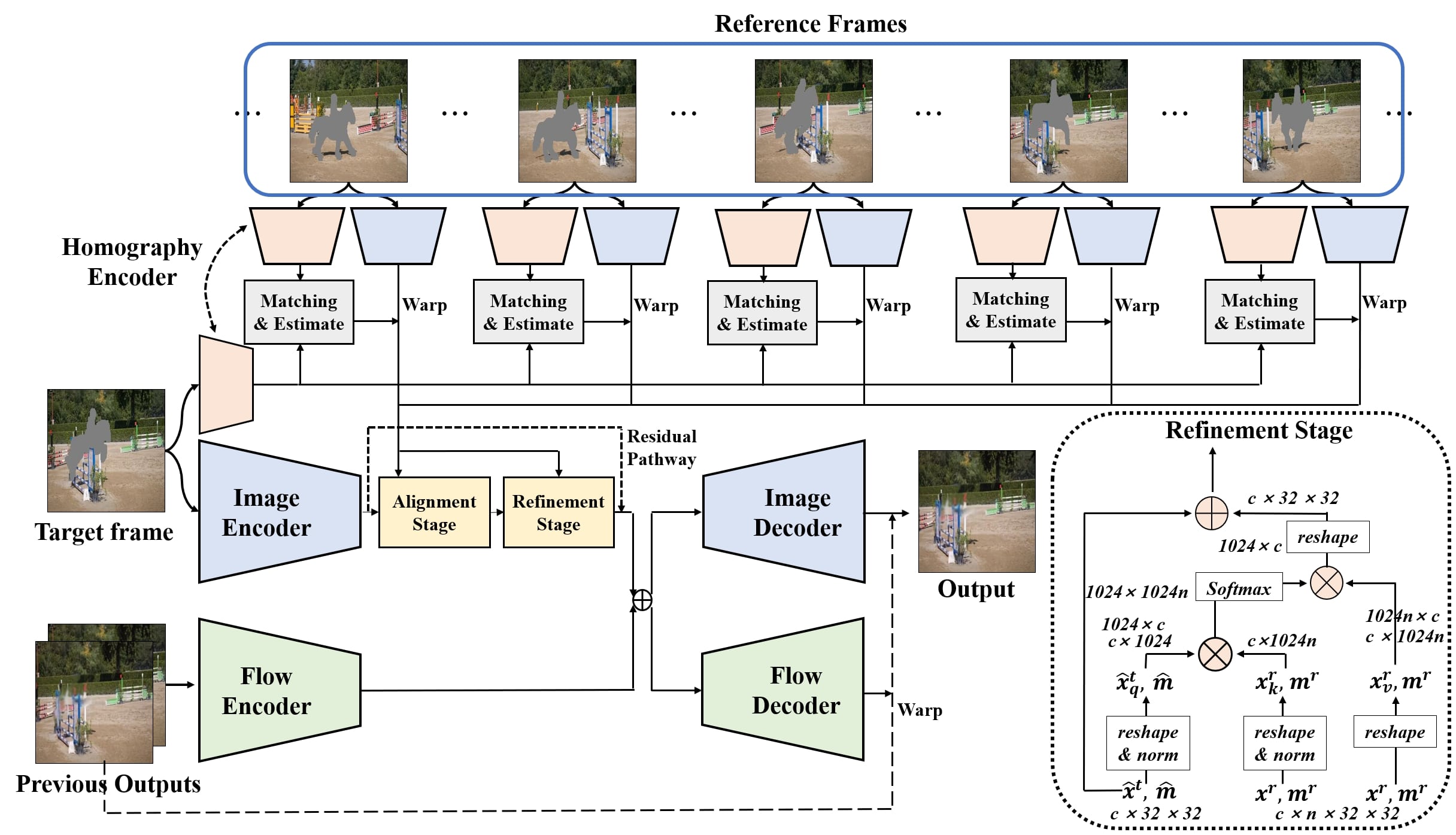}
\caption{\textbf{Network overvew.}}
\label{fig:network_overview}
\end{figure*}

\subsection{Network Design}

The overview of our network is shown in \figref{fig:network_overview}. The whole architecture can be divided into three parallel pathways: homography estimator, align-and-attend video inpainter, and flow estimator. 

\subsubsection{Homography estimator}
\label{homography}
Given a set of reference frames and a target frame, the goal of homography estimator is to produce transformation parameters $\theta$, which is to warp and align each reference frame onto the target frame. 

\textbf{Homography Encoder} takes an image of size $256 \times 256$ pixels as input, and produces an embedded feature map $f \in \mathbb{R}^{c \times 32 \times 32}$, where $c$ denotes the channel size . We use a same, shared encoder for both the reference and target frames. We denote features of any reference frame by $f^r$, and those of the target frame by $f^t$. 

\textbf{Masked matching} produces a measure of similarity between the reference and target feature maps. We denote the matching function as $M(f^{r}, f^{t}) := C$, such that $M:\mathbb{R}^{c\times32\times32} \times \mathbb{R}^{c\times32\times32} \rightarrow \mathbb{R}^{1024\times1024}$, where $C$ is a cosine similarity map computed between channel-wise normalized $f^r$ and $f^t$. We constrain the matching to happen only between the visible parts to deal with the holes regions. 
To this end, we use downsampled binary inpainting masks $m^r$ and $m^t \in \mathbb{R}^{32 \times 32}$. With $i$ and $j$ denoting the spatial grid indices for $f^r$ and $f^t$ respectively, the correlation map is computed as:

\begin{equation}
C(i,j) =
    \begin{cases}
        f^{r}(i)^{T}f^{t}(j) & \text{if}\ m^{r}(i)m^{t}(j) = 1 \\
        0 & \text{otherwise}
    \end{cases}
\label{eqn:our_formulation1}
\end{equation}

The similarity $C(i,j)$ is normalized by the softmax over the spatial dimension of $f^{r}$, for each $f^{t}(i)$.

\textbf{Transformation estimator} takes the correlation map $C$ as input and produces homography parameters $\theta$ between the reference and target. It is trained to output 6 parameters (\ie affine transformation), such that $R(C) := \theta$ , and $R:\mathbb{R}^{1024\times1024} \rightarrow \mathbb{R}^{6}$.

\subsubsection{Align-and-Attend Video Inpainter}
Our video inpainter is an encoder-decoder model consisting of following components that are designed to reconstruct the target holes in a coarse-to-fine manner.

\textbf{Image Encoder} part follows the same architecture as in the homography estimator. Similarly, we denote encoded features of any reference frame by $x^r$, and those of the target frame by $x^t$, both of spatial size $32 \times 32$ pixels.

\textbf{Alignment stage} is given the homography parameters computed as in~\secref{homography}, and accordingly aligns the reference feature maps onto the target feature map. We denote a reference feature map that is aligned to the target by $x^{r_{i} \rightarrow t}$, where $i \in [1...n]$, and $n$ denote frame index and the number of reference frames, respectively.  

After the alignment is to pick up the most relevant reference feature points in the spatial-temporal search window. We present an aggregation function that can evaluate the alignment for each reference feature maps, and exclude irrelevant information such as newly introduced scene parts. We measure the Euclidean distances (\ie L2-norm) between each aligned reference frames and the target frame, while ignoring the hole regions using the binary inpainting masks:

\begin{equation}
w^{r_{i}} = ||m^{t} m^{r_{i}\rightarrow t}(x^{t}-x^{r_{i}\rightarrow t})||_{2},
\label{eqn:our_formulation2}
\end{equation}

where smaller value of $w^{r_{i}}$ represents better alignment of $i$-th reference frame. The distance measure is used as a weighting coefficient, multiplied with corresponding inpainting masks $m^{r_i \rightarrow t}$. It is followed by softmax across temporal dimension to obtain a volume $A^{r_i}$ that weighs relevant pixels in the stack of $x^{r \rightarrow t}$ and flattens the stack into \textit{one-frame} feature map:

\begin{equation}
\hat{x}^{r} = \sum_{i} A^{r_{i}}x^{r_{i}\rightarrow t}, \quad A^{r_{i}} = \frac{ exp(\frac{m^{r_{i}\rightarrow t}}{w^{r_{i}}}) } { {\sum_{i}exp(\frac{m^{r_{i}\rightarrow t}}{w^{r_{i}}})} }
\label{eqn:our_formulation3}
\end{equation}

We identify the visible region that can be borrowed from with reference frames by $\hat{m}$, and the initial coarse prediction of the target feature map $\hat{x}^{t}$ is then obtained as:

\begin{equation}
\hat{x}^{r} = \sum_{i} A^{r_{i}}x^{r_{i}\rightarrow t}, \quad A^{r_{i}} = \frac{ exp(\frac{m^{r_{i}\rightarrow t}}{w^{r_{i}}}) } { {\sum_{j}exp(\frac{m^{r_{j}\rightarrow t}}{w^{r_{j}}})}}
\label{eqn:our_formulation4}
\end{equation}

\textbf{Refinement stage} is designed to model pixel-wise correspondences~\cite{wang2018non}, \eg, non-rigid motions, that cannot be covered by the previous global alignment stage. We propose to match the coarsely generated patches with the non-hole patches in the reference frame stack.  The pixel-wise matching between the reference and target feature maps is described in \figref{fig:network_overview}. The proposed non-local attention picks up the most relevant and best matching patches in the spatial-temporal search window, and aggregates them into the target hole regions to make a refined prediction. This module is designed to be non-parametric, not requiring any embedding layers.

\begin{equation}
\begin{split}
\hat{x}^{t}_{residual} = softmax((\hat{x}^{t}_{q})^{T}x^{r}_{k})(x^{r}_{v})^{T}, \\
\hat{x}^{t}_{refine} = 
    \begin{cases}
        \hat{x}^{t}_{residual} + \hat{x}^{t} & \text{if}\ m^{r}(i)\hat{m}(j) = 1 \\
        \hat{x}^{t} & \text{otherwise}
    \end{cases}
\label{eqn:our_formulation5}
\end{split}
\end{equation}

where $\hat{x}^{t}_{q}$, $\hat{x}^{r}_{k}$, $x^{r}_{v}$ are a feature reshaped into a matrix with a shape of $\mathbb{R}^{c\times1024}$, $\mathbb{R}^{c\times1024n}$, and $\mathbb{R}^{c\times1024n}$ respectively. Only $\hat{x}^{t}_{q}$ and $\hat{x}^{r}_{k}$ are L2-normalized over channel axis for attention map computation. $x^{t}_{residual}$ is properly reshaped before summation with $\hat{x}^{t}$.

\textbf{Residual pathway} is another convolutional pathway in parallel with the align-and-refine pathway. It is designed to allow the network to learn single image inpainting, which is to hallucinate novel contents that are never visible throughout the search window. The two pathways are aggregated and fed into single decoder to obtain the final output.

\textbf{Image Decoder} takes the aggregation of the two pathways, together with the warped features of the previous output frames $\hat{Y}_{t-2}^{t-1}$. In our preliminary experiment, we found that adding the intermediate representations from the previous time steps not only provides rich training signals to the whole network, but also enhances the temporal coherency in video results. The decoder recovers the fine details for the hole regions to generate raw output,  $\hat{Y'}_{t}$. It is designed with nearest-neighbor upsampling layers and following convolutions to prevent checkerboard artifacts.

\subsubsection{Optical flow estimator}
Our optical flow estimator is a simple encoder-decoder model.                                                                  It computes flow fields between the previous output frame and  the current target frame, that is used to enforce temporal consistency.

\textbf{Flow Encoder} takes previous two output frames as input in order to propagate reusable information to the current time step. The encoded features are also fed into the decoder of the video inpainter. 

\textbf{Flow Decoder} outputs optical flow from time step $t-1$ to $t$, and a composition mask.  We use the predicted flow $\hat{W}_{t\Rightarrow t-1}$ to warp the previous output $\hat{Y}_{t-1}$ onto the current time step $\hat{Y'}_{t}$. We then blend the two frames into one by the estimated composition mask $m'$ to obtain the final output of our whole network:

\begin{equation}
    \hat{Y}_{t} = (1-m') \odot \hat{Y'}_{t} + m' \odot \hat{W}_{t\Rightarrow t-1}(\hat{Y}_{t-1}).
\end{equation}

\subsection{Objective functions}

\textbf{Homography estimation.} We train the homography network using this objective function: 

\begin{equation}
\mathcal{L}_{align} =  (\frac{1}{n}\sum_{i=1}^{n}||G(\theta_{f^{t}}) - G(\theta_{f^{t^{*}}})||_{2}) + (||\theta_{f^{t}} - \theta_{f^{t^{*}}}||)
\label{eqn:our_formulation6}
\end{equation}

The first part is introduced by ~\cite{rocco2017convolutional}, and the second part is direct L1 loss of transformation parameters.
$G(\theta_{f^{t}})$ and $G(\theta_{f^{t^{*}}})$ correspond to the bilinear sampling grids that use predicted parameters and ground-truth respectively. Here, n denotes the total number of sampling coordinates.

Note that, the homography estimation network is trained independently from the video inpainting network (~\ie inpainting network and the optical flow estimation network). After the training, we freeze the network's parameters and use it as a global affine transformer.

\textbf{Video inpainting.} The objective function is designed to capture pixel-wise reconstruction accuracy, perceptual similarity, and temporal consistency.

The pixel-wise reconstruction loss is defined as follows:

\begin{equation}
\begin{split}
\mathcal{L}_{hole}  &= \sum_{t} \left\|(1-m^{t}) \odot (\hat{Y_t} - Y_t) \right\|_{1}, \\
\mathcal{L}_{valid} &= \sum_{t} \left\| m^{t} \odot (\hat{Y_t} - Y_t) \right\|_{1}
\end{split}
\end{equation}

where t indexes over the number of recurrences, $m_t$ is the binary mask, $\hat{Y_t}$ is the model output, $Y_t$ is the ground truth, and $\odot$ indicates the element-wise multiplication.

To ensure perceptual similarity between the predicted output and the ground truth, we adopt both image GAN loss, $\mathcal{L}_{im\_GAN}$, and video GAN loss, $\mathcal{L}_{vid\_GAN}$, that are introduced by \cite{wang2018video}:

For the temporal consistency, we use flow loss and warping loss which are defined as:

\begin{equation}
\begin{split}
\mathcal{L}_{flow} = \sum\limits_{t=2}^{T} (\left \| {W}_{t\Rightarrow t-1} - \hat{W}_{t\Rightarrow t-1} \right\|_{1} + \left \| {Y}_{t} - \hat{W}_{t\Rightarrow t-1}({Y}_{t-1}) \right\|_{1}), \\
\mathcal{L}_{warp} = \sum\limits_{t=2}^T M_{t \Rightarrow t-1} \left \| \hat{Y_t} - {W}_{t \Rightarrow t-1}({Y}_{t-1})\right\|_{1}
\end{split}
\end{equation}

where ${W}_{t\Rightarrow t-1}$ is the pseudo-groundtruth backward flow between the target frames, ${Y}_{t}$ and ${Y}_{t-1}$, extracted by FlowNet2~\cite{ilg2017flownet}, $M_{t \Rightarrow t-1}$ is the binary occlusion mask~\cite{lai2018learning}. Note that we use groundtruth target frames in the warping operation since the synthesizing ability is imperfect during training. We employ a curriculum learning scheme that increases the number of recursion, t, by 6 every 5 epochs. We increase the t up to 24.

The total loss is the weighted summation of all the loss functions:

\begin{equation}
\begin{split}
     \mathcal{L} &= 100\cdot\mathcal{L}_{hole} + 50\cdot\mathcal{L}_{valid} \\
                 &+ \mathcal{L}_{im\_GAN} + \mathcal{L}_{vid\_GAN} + 20\cdot\mathcal{L}_{flow}+20\cdot\mathcal{L}_{warp},
\end{split}
\label{eqn:total_loss}
\end{equation}

\subsection{Training}

\textbf{Homography estimation.}\quad We generate synthetic data using Places2 image dataset~\cite{zhou2018places}. Given a random image $I_{A}$, we generate the counterpart $I_{B}$ by applying an arbitrary transformation to $I_{A}$. This provides us great flexibility to gather as many training data as needed, for any 2D geometric transformation. To simulate diverse hole shapes and sizes, we use the irregular mask dataset~\cite{liu2018image} which consists of random streaks and holes of arbitrary shapes. During training, we apply random affine transformations (~\eg translation, rotation, scaling, sheering) to the mask. All images are resized to $256\times256$ pixels for training.

\textbf{Video inpainting.}\quad 
We employ a two-stage training scheme; 1) We first train the video inpainter without the alignment and the refinement stages to focus on learning a pure synthesis ability. To synthesize the training data, we follow the same protocol mentioned above. 2) We then add previously excluded stages along with the recurrence stream to the model. We fine-tune the whole model using videos in the Youtube-VOS dataset~\cite{xu2018youtube}.
It is a large-scale video segmentation dataset containing 4000+ YouTube videos with 70+ various moving objects. 
Since the most realistic appearance and motion can be obtained from the foreground segmentation masks, we use them to synthesize the training video data.
All video frames are resized to $256\times256$ pixels for training.

\subsection{Testing}

We use DAVIS dataset~\cite{perazzi2016benchmark, pont20172017}, which is widely used for video inpainting benchmarking. The videos are very challenging since they include dynamic scenes, complex camera movements, motion blur effects, and large occlusions. We obtain the inpainting mask by dilating the ground truth segmentation mask. Our method processes frame recursively in a sliding window manner.  

\subsection{Implementation Details}

Our model is implemented using Pytorch v0.4, CUDNN v7.0, CUDA v9.0. It run on the hardware with Intel(R) Xeon(R) (2.10GHz) CPU and NVIDIA GTX 1080 Ti GPU. The model runs at 15 fps on a GPU for frames of $256\times256$ pixels. We use Adam optimizer with $\beta$ = (0.9, 0.999). The learning rate starts with 2e-4 and divided by 10 every 5 epochs. We train our model from scatch. The homography training and video inpainting  training takes about 3 day each using eight NVIDIA GTX 1080 Ti GPUs.

\section{Experiments}

We evaluate our method both quantitatively and qualitatively. We compare out approach with state-of-the-art methods in three representative streams of study: deep image inpainting~\cite{yu2018generative}, deep video inpainting~\cite{kim2019video}, and optimization-based video inpainting~\cite{huang2016temporally}. Two metrics are mainly used for the evaluation. The first is the Inception score (FID)~\cite{wang2018video} extended to videos to measure the perceptual quality in spatio-temporal dimension. The second is flow warping errors between frames that measure temporal consistency of video results. We also conduct extensive ablation studies to validate the proposed design choices.

\begin{table}[t]
 \centering
   \subfloat[\scriptsize User study results.]{%
  \resizebox{.62\textwidth}{!}{%
        \begin{tabular}{c c}
        \hline
        \hline
        \multicolumn{2}{c}{Human Preference Score}  \\
        \hline
        Align-and-Attend Net (ours) / Huang~\etal~\cite{huang2016temporally} / Tie &  0.30 / 0.33 / 0.37 \\
        Align-and-Attend Net (ours) / Kim~\etal~\cite{kim2019video} / Tie & 0.58 / 0.19 / 0.23 \\
        \hline
        \hline
    \end{tabular}
    \label{table:user_study}
    }
  }
    \subfloat[\scriptsize Speed Comparison.]
     {%
       \resizebox{.32\textwidth}{!}{%
    \begin{tabular}[c]{ c |c }
        \hline
        \hline
        Methods & time \\
        \hline
        Huang~\etal~\cite{huang2016temporally} & several minutes per video \\
        Kim~\etal~\cite{kim2019video} & $\sim$ 12 fps \\
        Ours & $\sim$ 15 fps \\
        \hline
        \hline
    \end{tabular}
    \label{table:speed}
      }
  }
  \hspace{5mm}
      \subfloat[\scriptsize Spatial-Temporal Video Quality.]
     {%
       \resizebox{.35\textwidth}{!}{%
    \begin{tabular}[c]{ c |c }
        \hline
        \hline
        Methods & FID score \\
        \hline
        Yu~\etal~\cite{yu2018generative} & 9.6948 ($\pm$ 0.708)\\
        Huang~\etal~\cite{huang2016temporally} & 5.7032 ($\pm$ 1.158) \\
        Kim~\etal~\cite{kim2019video} & 6.2852 ($\pm$ 1.769) \\
        Ours & 5.775 ($\pm$ 1.707) \\
        \hline
        \hline
    \end{tabular}
    \label{table:sota_fid}
      }
  }
      \subfloat[\scriptsize Temporal consistency.]
     {%
       \resizebox{.35\textwidth}{!}{%
    \begin{tabular}[c]{ c |c }
        \hline
        \hline
        Methods & Warping error \\
        \hline
        Yu~\etal~\cite{yu2018generative} & 0.0037 ($\pm$ 0.0003)\\
        Huang~\etal~\cite{huang2016temporally} & 0.0017 ($\pm$ 0.0001) \\
        Kim~\etal~\cite{kim2019video} & 0.0019 ($\pm$ 0.0002) \\
        Ours & 0.0019 ($\pm$ 0.0001) \\
        \hline
        \hline
    \end{tabular}
    \label{table:sota_tc}
      }
  }
\captionsetup{font=footnotesize}
\vspace{3mm}
\caption{Quantitative comparison with state-of-the-art video inpainting methods~\cite{yu2018generative,huang2016temporally,kim2019video} on DAVIS.}
\label{table:compare_sota}
\end{table}

\begin{figure*}
\centering
\includegraphics[width=1\textwidth]{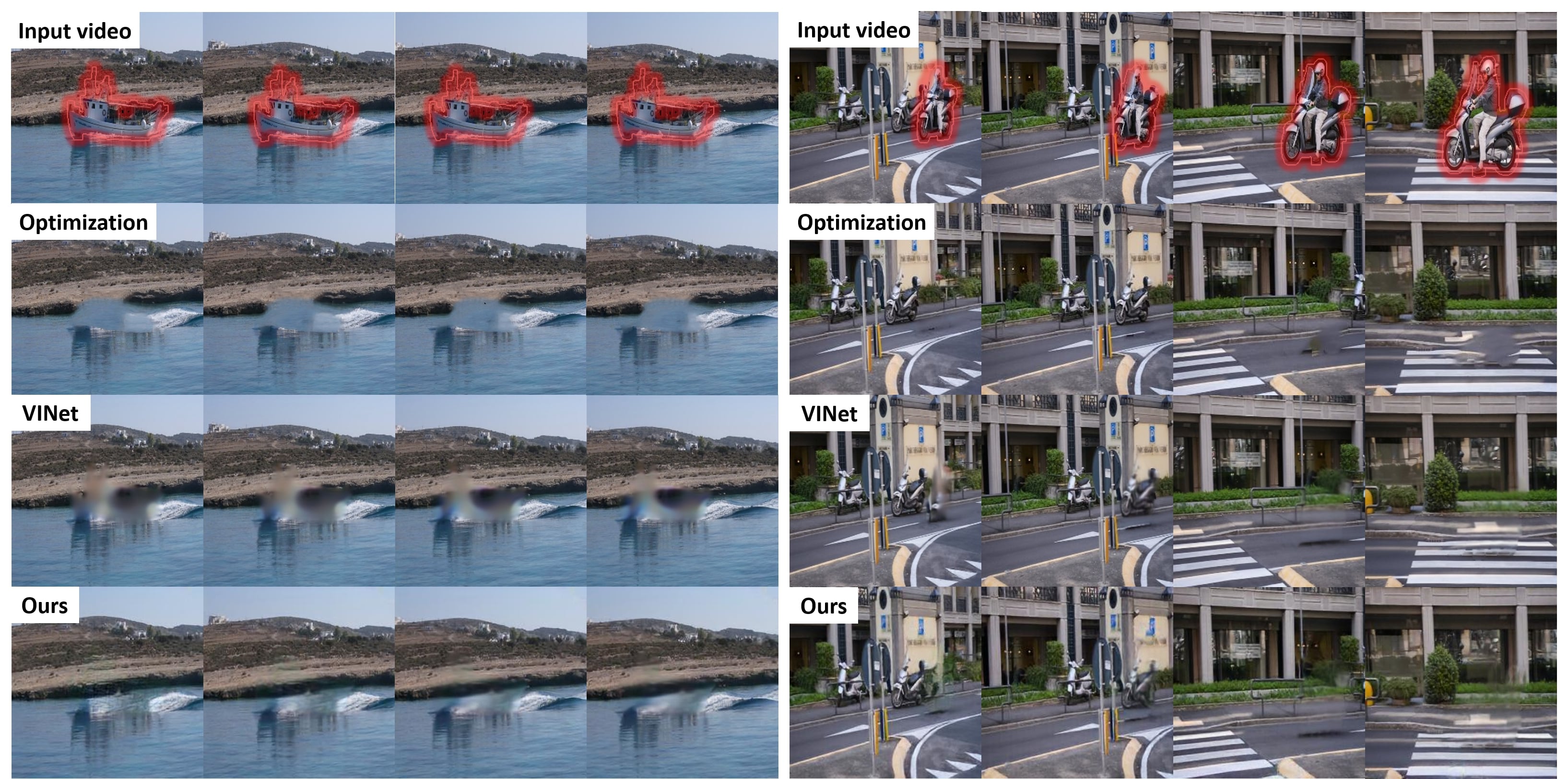}
\caption{Qualitative comparison with state-of-the-art video inpainting methods~\cite{huang2016temporally,kim2019video} on DAVIS.}
\label{fig:qual}
\end{figure*}

\subsection{User Study on Video Object Removal}

We perform a user study to evaluate the visual quality of inpainted videos. We use 20 videos from the DAVIS dataset and compare our method with the strong baselines~\cite{huang2016temporally,kim2019video}. A total of 25 users participated in this study. During each test, a user is shown video inpainting results by two different approaches, together with the input target video. We ask the user to check for both image quality and temporal coherency and to choose a better one. The users are allowed to play the videos multiple times to have enough time to distinguish the difference and make a careful judge. We report the ratio that each method outputs are preferred in~\tabref{table:user_study}. Our results are considered comparable to the~\cite{huang2016temporally}, and much higher-quality than~\cite{kim2019video} by the human subjects. Note that our method runs faster than both approaches (see \tabref{table:speed}). Some example results are shown in ~\figref{fig:qual}.

\subsection{Quantitative comparison}
\label{qaul_comp}

We further compare our method with the baselines~\cite{yu2018generative,huang2016temporally,kim2019video} using both FID score and warping loss. Since we need the ground truth videos for this experiment, we composite target videos by overlaying foreground mask sequences extracted from other videos. To measure the FID score, we take 20 videos in the DAVIS dataset. For each video, we ensure to choose a different video out of the other 19 videos to make a mask sequence. We use the first 64 frames of both input and mask videos. 
To measure the flow warping errors, we use Sintel dataset since it provides ground-truth optical flows. We take 32 frames each from 21 videos in Sintel dataset and randomly select 21 videos of length 32+ from DAVIS dataset to create corresponding mask sequences. 
For both metrics, we run five trials and average the scores over the videos and trials.
We summarize the results in \tabref{table:sota_fid} and \tabref{table:sota_tc}. We observe a similar tendency to the user study result.

\begin{table}[t]
 \centering
   \subfloat[\scriptsize Ablations on network design.]{%
  \resizebox{.42\textwidth}{!}{%
    \begin{tabular}{c c| c |c }
        \hline
        \hline
        \multicolumn{2}{c}{\textbf{Multi-frame aggregation}} &\multicolumn{1}{c}{\textbf{Output Propagation}} & \\
         Align  & Refine & Flow estimator & FID score    \\
        \hline
                       &           &             &8.966 ($\pm$0.709)     \\
        \hline
            \checkmark &           &             &8.139 ($\pm$1.017)     \\
                       &\checkmark &             &8.577 ($\pm$0.838)     \\
            \checkmark &\checkmark &             &8.262 ($\pm$1.615)     \\
        \hline
                       &           &\checkmark   &7.515 ($\pm$0.608)     \\
            \checkmark &           &\checkmark   &7.196 ($\pm$1.863)     \\
                       &\checkmark &\checkmark   &7.149 ($\pm$1.753)     \\
            \checkmark &\checkmark &\checkmark   &\textbf{5.775 ($\pm$1.707)}    \\
        \hline
        \hline
    \end{tabular}
    \label{table:Ablation_1}
    }
  }
    \subfloat[\scriptsize Ablations on non-local matching methods in the refinement stage.]
     {%
       \resizebox{.3\textwidth}{!}{%
    \begin{tabular}[c]{ c |c }
        \hline
        \hline
        Matching method & FID score \\
        \hline
        entire-entire & 7.156 ($\pm$1.818) \\
        hole-nonhole  & \textbf{5.775 ($\pm$1.707)} \\
        \hline
        \hline
    \end{tabular}
    \label{table:Ablation_2}
      }
  }
  \subfloat[\scriptsize Ablations on recurrence stream.]{%
        \resizebox{.28\textwidth}{!}{%
    \begin{tabular}[c]{ c |c }
        \hline
        \hline
        Flow estimator       & Warping error \\
        \hline
                      & 0.0027 ($\pm$0.0001) \\
        \checkmark    & \textbf{0.0019  ($\pm$0.0001)} \\
        \hline
        \hline
    \end{tabular}
    \label{table:Ablation_3}
    }
    }
\captionsetup{font=footnotesize}
\vspace{3mm}
\caption{Results of ablation studies.}
\label{table:Ablation}
\end{table}

\subsection{Ablation studies}

We run an extensive ablation study to demonstrate the effectiveness of different components of our method.
We measure FID score and warping error following the same protocol as in ~\secref{qaul_comp}.
The results are summarized in \tabref{table:Ablation}.

\textbf{Network design choices.} \quad
The main components of our network design are the two-stage feature aggregation part together with the temporal propagation part. First, we investigate the importance of each stages in the align-and-attend network. If we drop the alignment stage out of the pathway, the refinement stage alone has to pick up valid reference patches to fill in the holes. However, it is difficult for the non-local module to match the \textit{zero} patches to any reference patches without any priors. If we drop the refinement stage, the real video dynamics (\eg, small, non-rigid motions) cannot be modeled and the resulting videos would lack such fine details. To cancel out the effect of temporal propagation, we drop the flow estimator pathway. Without the recurrence, the temporal consistency is no longer well supported. If we remove multi-frame aggregation and the propagation parts, our network degenerates to a single image inpainting network. As shown in ~\tabref{table:Ablation_1}, all proposed components have complementary effects, and the best results are obtained when all components are fully used.


\textbf{Masked matching in non-local attention module.} \quad
In our non-local attention module, the coarsely completed region in the target hole is matched with the non-hole area in the reference frames. By doing so, the regions that still remain as the holes are ignored during the refinement matching; Only those newly generated patches are touched during the refinement stage. To see the effectiveness of this matching method, we show the results when there was no such constraint (~\ie entire patches in the target frame is matched with the entire patches in the reference frames). As shown in ~\tabref{table:Ablation_2}, we observe that our proposed matching method is indeed effective, resulting in better video quality.

\textbf{Recurrence stream.} \quad
We report the flow warping errors to compare the temporal consistencies of video results before and after adding the recurrence stream (flow estimator pathway).
As shown in ~\tabref{table:Ablation_3}, we observe the warping error is significantly reduced when there is the recurrence. This implies that propagating the previous output significantly improves the temporal consistency of videos. This is also consistent with the recent findings in~\cite{kim2019deep,kim2019video}.

\section{Conclusion}

In this paper, we present a novel deep network for video inpainting. Our model fills in a target hole by referring multiple reference frames in a coarse-to-fine manner. First, we propose homography-based alignment between the reference and target frames to roughly inpaint the missing contents. Second, a non-local attention module refines the previous generated regions. Both stages provide large spatial-temporal window size that have not been achieved by existing flow-based methods. We validate the effectiveness of our approach in real object removal scenarios.


{\small
\bibliographystyle{plain}
\bibliography{egbib}

\begin{thebibliography}{10}

\bibitem{ballester2001filling}
Coloma Ballester, Marcelo Bertalmio, Vicent Caselles, Guillermo Sapiro, and
  Joan Verdera.
\newblock Filling-in by joint interpolation of vector fields and gray levels.
\newblock In {\em {IEEE Trans. Image Processing (TIP)}}, volume~10, pages
  1200--1211. IEEE, 2001.

\bibitem{bertalmio2000image}
Marcelo Bertalmio, Guillermo Sapiro, Vincent Caselles, and Coloma Ballester.
\newblock Image inpainting.
\newblock In {\em Proceedings of the 27th annual conference on Computer
  graphics and interactive techniques}, pages 417--424, 2000.

\bibitem{bertalmio2003simultaneous}
Marcelo Bertalmio, Luminita Vese, Guillermo Sapiro, and Stanley Osher.
\newblock Simultaneous structure and texture image inpainting.
\newblock In {\em {IEEE Trans. Image Processing (TIP)}}, volume~12, pages
  882--889. IEEE, 2003.

\bibitem{darabi2012image}
Soheil Darabi, Eli Shechtman, Connelly Barnes, Dan~B Goldman, and Pradeep Sen.
\newblock Image melding: Combining inconsistent images using patch-based
  synthesis.
\newblock In {\em {ACM Trans. on Graph. (ToG)}}, volume~31, pages 82--1.
  Citeseer, 2012.

\bibitem{efros2001image}
Alexei~A Efros and William~T Freeman.
\newblock Image quilting for texture synthesis and transfer.
\newblock In {\em Proceedings of the 28th annual conference on Computer
  graphics and interactive techniques}, pages 341--346. ACM, 2001.

\bibitem{granados2012background}
Miguel Granados, Kwang~In Kim, James Tompkin, Jan Kautz, and Christian
  Theobalt.
\newblock Background inpainting for videos with dynamic objects and a
  free-moving camera.
\newblock In {\em {Proc. of European Conf. on Computer Vision (ECCV)}}, pages
  682--695. Springer, 2012.

\bibitem{granados2012not}
Miguel Granados, James Tompkin, K~Kim, Oliver Grau, Jan Kautz, and Christian
  Theobalt.
\newblock How not to be seen—object removal from videos of crowded scenes.
\newblock In {\em Computer Graphics Forum}, volume~31, pages 219--228. Wiley
  Online Library, 2012.

\bibitem{huang2016temporally}
Jia-Bin Huang, Sing~Bing Kang, Narendra Ahuja, and Johannes Kopf.
\newblock Temporally coherent completion of dynamic video.
\newblock {\em ACM Transactions on Graphics (TOG)}, 35(6):196, 2016.

\bibitem{iizuka2017globally}
Satoshi Iizuka, Edgar Simo-Serra, and Hiroshi Ishikawa.
\newblock Globally and locally consistent image completion.
\newblock {\em ACM Transactions on Graphics (TOG)}, 36(4):107, 2017.

\bibitem{ilg2017flownet}
Eddy Ilg, Nikolaus Mayer, Tonmoy Saikia, Margret Keuper, Alexey Dosovitskiy,
  and Thomas Brox.
\newblock Flownet 2.0: Evolution of optical flow estimation with deep networks.
\newblock In {\em {Proc. of Computer Vision and Pattern Recognition (CVPR)}},
  pages 2462--2470, 2017.

\bibitem{kim2019deep}
Dahun Kim, Sanghyun Woo, Joon-Young Lee, and In~So Kweon.
\newblock Deep blind video decaptioning by temporal aggregation and recurrence.
\newblock In {\em {Proc. of Computer Vision and Pattern Recognition (CVPR)}},
  2019.

\bibitem{kim2019video}
Dahun Kim, Sanghyun Woo, Joon-Young Lee, and In~So Kweon.
\newblock Deep video inapinting.
\newblock In {\em {Proc. of Computer Vision and Pattern Recognition (CVPR)}},
  2019.

\bibitem{lai2018learning}
Wei-Sheng Lai, Jia-Bin Huang, Oliver Wang, Eli Shechtman, Ersin Yumer, and
  Ming-Hsuan Yang.
\newblock Learning blind video temporal consistency.
\newblock In {\em {Proc. of European Conf. on Computer Vision (ECCV)}}, 2018.

\bibitem{lee2019instert}
Donghoon Lee, Tomas Pfister, and Ming-Hsuan Yang.
\newblock Inserting videos into videos.
\newblock In {\em {Proc. of Computer Vision and Pattern Recognition (CVPR)}},
  2019.

\bibitem{levin2003learning}
Anat Levin, Assaf Zomet, and Yair Weiss.
\newblock Learning how to inpaint from global image statistics.
\newblock In {\em {Proc. of Int'l Conf. on Computer Vision (ICCV)}}, page 305.
  IEEE, 2003.

\bibitem{liu2018image}
Guilin Liu, Fitsum~A Reda, Kevin~J Shih, Ting-Chun Wang, Andrew Tao, and Bryan
  Catanzaro.
\newblock Image inpainting for irregular holes using partial convolutions.
\newblock In {\em {Proc. of European Conf. on Computer Vision (ECCV)}}, 2018.

\bibitem{newson2014video}
Alasdair Newson, Andr{\'e}s Almansa, Matthieu Fradet, Yann Gousseau, and
  Patrick P{\'e}rez.
\newblock Video inpainting of complex scenes.
\newblock {\em SIAM Journal on Imaging Sciences}, 7(4):1993--2019, 2014.

\bibitem{pathak2016context}
Deepak Pathak, Philipp Krahenbuhl, Jeff Donahue, Trevor Darrell, and Alexei~A
  Efros.
\newblock Context encoders: Feature learning by inpainting.
\newblock In {\em {Proc. of Computer Vision and Pattern Recognition (CVPR)}},
  pages 2536--2544, 2016.

\bibitem{perazzi2016benchmark}
Federico Perazzi, Jordi Pont-Tuset, Brian McWilliams, Luc Van~Gool, Markus
  Gross, and Alexander Sorkine-Hornung.
\newblock A benchmark dataset and evaluation methodology for video object
  segmentation.
\newblock In {\em {Proc. of Computer Vision and Pattern Recognition (CVPR)}},
  pages 724--732, 2016.

\bibitem{pont20172017}
Jordi Pont-Tuset, Federico Perazzi, Sergi Caelles, Pablo Arbel{\'a}ez, Alex
  Sorkine-Hornung, and Luc Van~Gool.
\newblock The 2017 davis challenge on video object segmentation.
\newblock {\em arXiv preprint arXiv:1704.00675}, 2017.

\bibitem{rocco2017convolutional}
Ignacio Rocco, Relja Arandjelovic, and Josef Sivic.
\newblock Convolutional neural network architecture for geometric matching.
\newblock In {\em {Proc. of Computer Vision and Pattern Recognition (CVPR)}},
  2017.

\bibitem{simakov2008summarizing}
Denis Simakov, Yaron Caspi, Eli Shechtman, and Michal Irani.
\newblock Summarizing visual data using bidirectional similarity.
\newblock In {\em {Proc. of Computer Vision and Pattern Recognition (CVPR)}},
  pages 1--8. IEEE, 2008.

\bibitem{wang2018video2}
Chuan Wang, Haibin Huang, Xiaoguang Han, and Jue Wang.
\newblock Video inpainting by jointly learning temporal structure and spatial
  details.
\newblock In {\em {Proc. of AAAI Conference on Artificial Intelligence
  (AAAI)}}, 2018.

\bibitem{wang2018video}
Ting-Chun Wang, Ming-Yu Liu, Jun-Yan Zhu, Guilin Liu, Andrew Tao, Jan Kautz,
  and Bryan Catanzaro.
\newblock Video-to-video synthesis.
\newblock In {\em {Proc. of Neural Information Processing Systems (NeurIPS)}},
  2018.

\bibitem{wang2018non}
Xiaolong Wang, Ross Girshick, Abhinav Gupta, and Kaiming He.
\newblock Non-local neural networks.
\newblock In {\em {Proc. of Computer Vision and Pattern Recognition (CVPR)}},
  pages 7794--7803, 2018.

\bibitem{wexler2004space}
Yonatan Wexler, Eli Shechtman, and Michal Irani.
\newblock Space-time video completion.
\newblock In {\em {IEEE Trans. Pattern Anal. Mach. Intell. (TPAMI)}}, pages
  120--127. IEEE, 2004.

\bibitem{xu2018youtube}
Ning Xu, Linjie Yang, Yuchen Fan, Jianchao Yang, Dingcheng Yue, Yuchen Liang,
  Brian Price, Scott Cohen, and Thomas Huang.
\newblock Youtube-vos: Sequence-to-sequence video object segmentation.
\newblock In {\em {Proc. of European Conf. on Computer Vision (ECCV)}}, 2018.

\bibitem{yu2018free}
Jiahui Yu, Zhe Lin, Jimei Yang, Xiaohui Shen, Xin Lu, and Thomas~S Huang.
\newblock Free-form image inpainting with gated convolution.
\newblock {\em arXiv preprint arXiv:1806.03589}, 2018.

\bibitem{yu2018generative}
Jiahui Yu, Zhe Lin, Jimei Yang, Xiaohui Shen, Xin Lu, and Thomas~S Huang.
\newblock Generative image inpainting with contextual attention.
\newblock In {\em {Proc. of Computer Vision and Pattern Recognition (CVPR)}},
  2018.

\bibitem{zhou2018places}
Bolei Zhou, Agata Lapedriza, Aditya Khosla, Aude Oliva, and Antonio Torralba.
\newblock Places: A 10 million image database for scene recognition.
\newblock In {\em {IEEE Trans. Pattern Anal. Mach. Intell. (TPAMI)}},
  volume~40, pages 1452--1464. IEEE, 2018.

\end{thebibliography}
}
\end{document}